# Personalized Saliency and Its Prediction

Yanyu Xu *, Shenghua Gao *, Junru Wu *, Nianyi Li, and Jingyi Yu

**Abstract**—Nearly all existing visual saliency models by far have focused on predicting a universal saliency map across all observers. Yet psychology studies suggest that visual attention of different observers can vary significantly under specific circumstances, especially a scene is composed of multiple salient objects. To study such heterogenous visual attention pattern across observers, we first construct a personalized saliency dataset and explore correlations between visual attention, personal preferences, and image contents. Specifically, we propose to decompose a personalized saliency map (referred to as PSM) into a universal saliency map (referred to as USM) predictable by existing saliency detection models and a new discrepancy map across users that characterizes personalized saliency. We then present two solutions towards predicting such discrepancy maps, *i.e.*, a multi-task convolutional neural network (CNN) framework and an extended CNN with Person-specific Information Encoded Filters (CNN-PIEF). Extensive experimental results demonstrate the effectiveness of our models for PSM prediction as well their generalization capability for unseen observers.

✦

## 1 INTRODUCTION

SALIENCY refers to a component (object, pixel, person) in a scene that stands out relative to its neighbors and the problem lies at the center of human perception and cognition. Traditional saliency detection techniques attempt to extract the most pertinent subset of the captured sensory data (RGB images or light fields) for predicting human visual attention. Applications are numerous, ranging from compression [1] to image re-targeting [2], and most recently to virtual reality and augmented reality [3].

By far, almost all previous approaches have focused on exploring a universal saliency model, *i.e.*, to predict potential salient regions common to observers while ignoring their differences in gender, age, personal preferences, *etc.*. Such universal solutions are beneficial in the sense there are able to capture all "potential" saliency regions. Yet they are insufficient in recognizing heterogeneity across individuals. Examples in Fig. 1 illustrate that while multiple objects are deemed highly salient within the same image *e.g.*, *human face* (first row), *text* (last two rows, '*DROP COFFEE*' in the last row) and objects of *high color contrast* (zip-top cans in the third row), different observers have very different fixation preferences when viewing the image. For the rest of the paper, we use the term *universal saliency* to describe salient regions that incur high fixations across all observers via a universal saliency map (referred to as USM); in contrast, we use the term *personalized saliency* to describe the heterogeneous ones via a personalized saliency map (referred to as PSM).

Heterogeneity in saliency preference has been widely recognized in psychology: "Interestingness is highly subjective and there are individuals who did not consider any image interesting in some sequence" [4]. Such inconsistency in saliency preference plays a critical role in perception and recognition [34] [29] [30]. In fact, if one knows an observer's personalized interestingness (personalized saliency) for a scene, one can potentially design tailored algorithms to cater to his/her needs. Personalized saliency detection hence can potentially benefit various applications. For

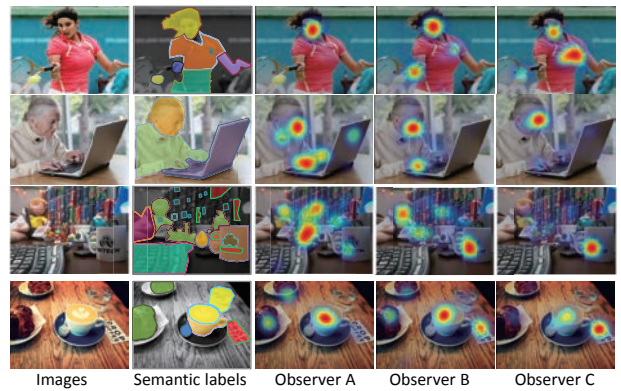

Images    Semantic labels    Observer A    Observer B    Observer C

Fig. 1: An illustration of our PSM dataset. Our dataset provides both eye fixations of different subjects and semantic labels of all images. Due to a large number of objects in our dataset, for each image, instead of fully segmenting every object, we only label objects that cover at least three gaze points from each individual. To reliably obtain the PSM, we have each subject to view the image 4 times. The commonality across different observers is characterized by the USM whereas the discreminility is characterized by the PSM.

example, in image retargeting, texts on the table in the fourth row in Fig. 1 should be preserved for observer B and C when resizing the image but can be eliminated for observer A. In VR content streaming, we can realize user-correlative compression algorithm: we can design data compression algorithms that preserve personalized salient regions but further reduce the rest to minimize transmission overhead. Finally, in advertisement deployment, we can adapt the location of the advertising window according to the predicted personal preferences.

Despite its usefulness, very little work has focused on directly characterizing personalized saliency: it is challenging to gather sufficient gaze information and there is little industry support to construct a comprehensive dataset suitable for personalized saliency detection. In this paper, we first construct the first personalized

• *Yanyu Xu, Junru Wu, Shenghua Gao and Jingyi Yu are with ShanghaiTech University, China. Nianyi Li is with University of Delaware, United States.*

saliency detection dataset that consists of 1600 images viewed by 30 observers. To improve reliability, each image is viewed by each observer 4 times at different times within a week. We use an '*Eyefollower*' eye tracker to record each individual's gaze patterns and produce a set of 48,000 ($1,600 \times 30$) eye fixation maps. The total time for data collection takes over 255 hours. To correlate the acquired PSMs and the image contents, we manually segment each image into a collection of objects and semantically label them. Examples in Fig. 1 illustrate how fixations vary across three different subjects. Our annotated dataset provides fine-grained semantic analysis for studying saliency variations across individuals. For example, we observe that certain types of objects such as watches, belts would introduce more incongruity (due to gender differences) whereas other types of objects such as faces lead to more coherent fixation maps, as shown in Fig. 2.

In this paper, we present a computational model towards this personalized saliency detection problem. As shown in Fig. 1, saliency maps from different observers still share certain commonality, which is encoded in USM. Hence, we propose to model the PSM as a combination of USM and a discrepancy map which is related to each observer's personal preferences and the image contents. To model this discrepancy map, we propose two solutions. In the first solution, we adopt a multi-task convolutional neural network (CNN) framework to identify the discrepancy between PSM and USM for each person, as shown in Fig. 4. In the second solution, motivated by the findings in psychology [36] [34] [29] [30] that such discrepancy is closely related to the observer's personal information, including gender, age, personal preferences, *etc.*, we propose to encode the personal information into the CNN filters for discrepancy prediction, and term such solution as CNN with Person-specific Information Encoded Filters (CNN-PIEF), as shown in Fig. 5.

The rest of this paper is organized as follows: In Section 2, we review related methods and datasets used in classic universal saliency detection. In Section 3, we describe our personalized saliency detection dataset. In Section 4, we present our solutions for personalized saliency detection. Experimental evaluations are described in Section 5 and we discuss limitations and future work in Section 6.

## 2 RELATED WORK

Tremendous efforts on saliency detection have been focused on predicting universal saliency. In this section, we will briefly discuss the most relevant ones, including both hand-crafted features based and deep learning based saliency detection methods. We refer the readers to [5] for a comprehensive study on existing universal saliency detection schemes and we only discuss work most related to the proposed personalized saliency.

### 2.1 Universal Saliency Detection

#### 2.1.1 Universal Saliency Detection Benchmarks.

There are a few widely used saliency object detection and fixation prediction datasets, in which each image is generally associated with a single ground truth saliency map, averaged across the fixation maps across the participates. To select images suitable for personalized saliency, we explore several popular eye fixation datasets. The MIT dataset [6] contains 1,003 images viewed by 15 subjects. In addition, the PASCAL-S [7] dataset provides the ground truth for both eye fixation and object detection and consists of 850 images viewed by 8 subjects. The iSUN dataset [52], a large scale dataset used for eye fixation prediction, contains 20,608 images from the SUN database. The images are completely annotated and are viewed by users. Finally, the SALICON dataset [9] consists of 10,000 images with rich contextual information.

#### 2.1.2 Hand-crafted Features Based Saliency Detection.

The work of [46], one of the classic saliency models, used the predefined color subspaces such as intensity, red-green, and blue-yellow color opponencies and four orientations to represent the image and simulate the receptive fields of various neurons through local center-surround differences to predict saliency maps. [47] employed an intra and inter channel fusion strategy for color and orientation feature channels with multi-scale rarities to predict saliency map. [48] simulated the saliency through calculating the distance between low-level features including color, orientation, and spatial features extracted from a local image patch. [50] used high-level features extracted from person and face areas to address this saliency task. [49] investigated the effect of high-level features including the presence of text and cars in estimating saliency maps.

#### 2.1.3 CNN Based Saliency Detection.

Our personalized saliency detection exploits convolutional neural network (CNN) in light of its great success in multiple computer vision tasks, *e.g.*, image classification [40], semantic segmentation [41], as well as saliency detection [42] [43] [12] [13]. An early approach of Ensembles of Deep Networks (eDN) [42] was proposed by Vig *et al.*, where feature maps from different layers in a 3-layers ConvNet are fed into a simple linear classifier for salient or non-salient classification. Later a DeepVisual attention model proposed by Kummerer *et al.* [43], leveraged the AlexNet [40] trained for image classification to extract features for eye fixation regression. In a similar vein, Huang *et al.* [9] proposed to fine-tune CNNs pre-trained for object recognition (AlexNet [40], VGG-16 [44] and GoogLeNet [45]) via a new objective function based on the saliency evaluation metrics, *e.g.*, Normalized Scanpath Saliency (NSS), Similarity, and KL-Divergence. Pan *et al.* [10] proposed to use a shallow CNN trained from scratch and another deep CNN where the weights of its first 3 layers are adapted from VGG_CNN_M trained for image classification for saliency map regression. Liu *et al.* [27] proposed a multi-resolution CNN where three final fully connected layers are combined to form the final saliency map. Srinivas *et al.* presented a DeepFix [12] network by using Location Biased Convolution filters to allow the network to exploit location dependent patterns. Kruthiventi *et al.* [13] proposed a unified framework to predict eye fixation and segment salient objects. All these approaches have focused on the universal saliency models and we show many merits of these techniques can be used for personalized saliency.

### 2.2 Personalized saliency

Recently, the heterogeneity of saliency maps across different subjects has attracted the attention of researchers in computer vision community. Specifically, in [53], Krishna *et al.* investigated the effect of gaze on age allocation for observers of different ages. However, we consider more factors that related to visual attention including gender, major, personal preference, *etc.*. In [54], Jiang *et al.* used the differences in eye movements between healthy people and those with Autism Spectrum Disorder to classify clinical populations, while our work focuses on the prediction of personalized saliency maps. In [55], Kummerer *et al.* studied the relevance of low- versus high-level features in predicting fixation locations, while we propose to predict personalized saliency maps for each observer based his/her personal information.

| | Person 1 | Person 2 | Person 3 | Person 4 | Person 5 | Person 6 | Person 7 |
|---|---|---|---|---|---|---|---|
| men bow tie | 0.068388 | 0.046459 | 0.035015 | 0.07911 | 0.025138 | 0.027694 | 0.0458365 |
| women bow tie | 0.014818 | 0.019792 | 0.078912 | 0.109668 | 0.004215 | 0.023556 | 0.0969675 |
| men hand watch | 0.034834 | 0.034573 | 0.057979 | 0.036348 | 0.027059 | 0.085353 | 0.0374112 |
| women hand watch | 0.035535 | 0.04356 | 0.041277 | 0.033336 | 0.022686 | 0.087908 | 0.0246781 |
| violin | 0.01083 | 0.00374 | 0.046803 | 0.046346 | 0.014374 | 0.091471 | 0.0157963 |
| keyboard | 0.038245 | 0.046073 | 0.029626 | 0.060052 | 0.026649 | 0.049733 | 0.0286243 |
| oven | 0.018754 | 0.054371 | 0.084905 | 0.02281.4 | 0.061014 | 0.099879 | 0.0140854 |
| fish tank | 0.036864 | 0.041231 | 0.017544 | 0.026256 | 0.017181 | 0.027864 | 0.120358 |
| can | 0.033927 | 0.043147 | 0.035909 | 0.042654 | 0.02892 | 0.010673 | 0.0444906 |
| men face | 0.025989 | 0.04491 | 0.0429 | 0.03387 | 0.03736 | 0.037919 | 0.0384792 |
| women face | 0.027088 | 0.040768 | 0.043192 | 0.037849 | 0.035902 | 0.033328 | 0.0401711 |

Fig. 2: The distribution of the interestingness of various objects for the same participant described by the technique in Section 4.2.2. Higher value indicates that the participant pays more attention on the object.

# 3 PERSONALIZED SALIENCY MAP (PSM) DATASET

We start with constructing a dataset suitable for personalized saliency analysis. The dataset has been released to the computer vision community [1].

## 3.1 Data Collection

Clearly, the rule of thumb for preparing such a dataset is to choose images that yield distinctive fixation map among different persons. To do so, we first analyze existing datasets. A majority of existing eye fixation datasets provide the one-time visual attention tracking results of each individual human observer. Specifically, we can correlate the level of agreement across different observers with respect to the number of object categories in the image. When an image contains few objects, we observe that an observer tends to fix his/her visual attention at objects that have specific semantic meanings, *e.g.*, faces, texts, signs [6], [14]. These objects indeed attract more attention and hence be deemed more salient. However, when an image consists of multiple objects all with strong saliency as shown in Fig. 1, we observe an observer tends to diverge his/her attention. In fact, the observer focuses attention on objects that attract him/her most personally.

We therefore deliberately choose 1,600 images with multiple semantic annotations to construct our dataset for PSM purpose. Among them, 1,100 images are chosen from existing saliency detection datasets including SALICON [15], ImageNet [16], iSUN [52], OSIE [14], PASCAL-S [7], 125 images are captured by ourselves, and 375 images are gathered from the Internet.

Quantitative criteria are not used to guide the image selection. However, we compare the inter-subject consistency scores in both original dataset and our dataset in Table 1. It shows that our picked images are with small inter-subject consistency in both original dataset and our PSM dataset.

## 3.2 Ground Truth Annotations

To gather the ground truth, we recruit 30 student participants (14 males, 16 females, aged between 20 and 25). All participants have normal or corrected-to-normal vision. In our setup, each observer sits about 40 inches in front of a 24-inches LCD monitor with resolution of $1920 \times 1080$. All images are resized to the same resolution. We conduct all experiments in an empty and semi-dark room, with only one standby assistant. An eye tracker ('*Eyefollower*' eye tracker) records their visual attention as they view each image for 3 seconds. We partition 1,600 images into 34 sessions, each containing 40 to 55 images. Each session lasts about 3 minutes followed by a half minute break. The eye tracker is re-calibrated at the beginning of each session. To ensure the veracity of the fixation map of each individual as well as to remove outliers, we have each image be viewed by each observer 4 times. We then average the 4 saliency maps of the same image viewed by the given observer and use the result as the ground truth PSM of the image for the same observer. To obtain a continuous saliency map of an image from the raw data recorded by the eye tracker, we follow [6] by smoothing the fixation locations via Gaussian blurs.

| datasets | Observers | CC | Similarity | AUC-Judd | NSS |
|---|---|---|---|---|---|
| MIT | Fixations in MIT | 0.5308 | 0.4585 | 0.8934 | 2.1442 |
| MIT(subset) | Fixations in MIT | 0.4231 | 0.3981 | 0.8548 | 1.5643 |
| MIT(subset) | Fixations in ours | 0.4383 | 0.4215 | 0.8687 | 1.6238 |
| OSIE | Fixations in OSIE | 0.5139 | 0.4322 | 0.8878 | 2.7745 |
| OSIE(subset) | Fixations in OSIE | 0.4592 | 0.4095 | 0.8657 | 2.2523 |
| OSIE(subset) | Fixations in ours | 0.4389 | 0.4170 | 0.8538 | 1.7287 |
| PASCAL-S | Fixations in PASCAL-S | 0.4681 | 0.4154 | 0.8810 | 2.0563 |
| PASCAL-S(subset) | Fixations in PASCAL-S | 0.4614 | 0.4092 | 0.8629 | 1.8398 |
| PASCAL-S(subset) | Fixations in ours | 0.3746 | 0.3788 | 0.8365 | 1.4808 |

TABLE 1: The inter-subject consistency scores in the original dataset and those in our new dataset. The subset in bracket indicates the measurements are computed based on a subset of images picked from those datasets.

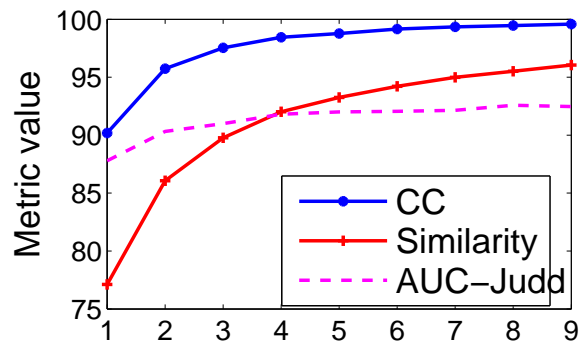

Fig. 3: The point at $x = n$ measures the differences between ground truth saliency maps generated by viewing the same image $n$ times and $n + 1$ times. This figure shows that when $n \geq 4$, the ground truth saliency maps generated by viewing the image $n$ times have little differences compared with that generated by observing the image $n + 1$ times. Thus viewing each image 4 times is enough to get a robust estimation of the PSM ground truth.

To further analyze the causes of saliency heterogeneity, we conduct the semantic segmentation for all 1,600 images via the open annotation tool LabelMe [17]. Specifically, we annotate 26,100 objects of 242 classes in total and identify objects that attract more attention for each individual participant. To achieve this, we compare the fixation map with the mask of a specific object and use the result as the attention value of the corresponding object. We then average the result over all images that containing the same object, and use it to measure the interestingness of the object to a specific participant. In Fig. 2, we illustrate some representative objects and persons and show the distribution of the interestingness of various objects for different participants. We observe that all participants exhibit a similar level of interestingness in faces where they exhibit different interestingness in various objects, such as watch, bow tie, *etc.*. This validates that it is necessary to choose images with multiple objects to build our PSM data.

|              | CC     | Similarity | AUC-Judd | NSS    |
|--------------|--------|------------|----------|--------|
| The $1^{st}$ time | 0.3504 | 0.3228     | 0.8427   | 1.0287 |
| The $2^{nd}$ time | 0.3447 | 0.3198     | 0.8387   | 1.0090 |
| The $3^{rd}$ time | 0.3435 | 0.3191     | 0.8384   | 1.0098 |
| The $4^{th}$ time | 0.3438 | 0.3192     | 0.8363   | 1.0096 |

TABLE 2: Center preference of saliency maps viewed at different time, evaluated with the method proposed in [53].

|              | The $1^{st}$ time | The $2^{nd}$ time | The $3^{rd}$ time | The $4^{th}$ time |
|--------------|-----------|-----------|-----------|-----------|
| Numbers      | 9.1667    | 8.8000    | 8.9667    | 9.1667    |
| Durations (s)| 0.2505    | 0.2506    | 0.2593    | 0.2548    |
| Distance     | 0.2961    | 0.3020    | 0.3032    | 0.3052    |

TABLE 3: The eye fixation distributions in our dataset

|                          | CC     | Similarity | AUC-Judd | NSS    |
|--------------------------|--------|------------|----------|--------|
| within-subject consistency | 0.5555 | 0.4948     | 0.8775   | 2.3038 |
| inter-subject consistency  | 0.4456 | 0.4144     | 0.8098   | 1.7843 |

TABLE 4: Inter-subject consistency vs. within-subject consistency in our dataset.

| Datasets | CC     | Similarity | AUC-Judd | NSS    |
|----------|--------|------------|----------|--------|
| MIT      | 0.5308 | 0.4585     | 0.8934   | 2.1442 |
| OSIE     | 0.5139 | 0.4322     | 0.8878   | 2.7745 |
| PASCAL-S | 0.4681 | 0.4154     | 0.8810   | 2.0563 |
| Ours     | **0.4456** | **0.4144** | **0.8098** | **1.7843** |

TABLE 5: Inter-subject consistency in different datasets. To compute the inter-subject consistency, we compute CC, Similarity, AUC-Judd, and NSS for pair-wise saliency maps viewed by different observers for each image, then we average the results over all images. For fair comparison, CC, Similarity, AUC-Judd, and NSS of our method reported here is based on the saliency maps viewed by each observer once.

### 3.3 Dataset Analysis

Our dataset is more suitable for personalized saliency than the existing ones for several reasons. We first show that it is necessary for a participant to view the same image multiple times. We further demonstrate that heterogeneity in saliency maps can be severe. Finally, we explain why other existing datasets are less useful for personalized saliency.

#### 3.3.1 Multiple vs. Single Viewing.

To validate whether it is necessary for an observer to view each image multiple times, we randomly sample 220 images, and each image is viewed by the same participant 10 times. The time interval for the same person to view the same image ranges from one day to one week because we want to get the short-term memory of the person for the given image.

We then calculate the differences of these saliency maps in terms of the commonly used metrics for saliency detection [18]: CC, Similarity, and AUC-Judd. We average these criteria for all persons and all images, and show the results in Fig. 3. We observe that the saliency map obtained by viewing each image only once vs. multiple times exhibit significant differences. Further, the saliency map, averaged over 4 is closer to the long-term result.

In order to investigate the differences between early and late experimental sessions, we calculate the average number of fixations, average fixation durations and fixation distributions, and list the results in Table. 3. We use the mean distance between each fixation and image center, normalized by the length of image diagonal to measure the fixation distribution. Further, following the calculation of center preference in [53], we first calculate the average saliency map across all images for each time, *i.e.*, the center map. We then use this center map to measure agreement scores with fixation locations for all images viewed at different time, and the images are shown in Table 2. Table 3 and Table 2 show that there is no significant difference for eye fixations viewed at different time.

#### 3.3.2 Heterogeneity Among Different Datasets.

To further illustrate that our proposed dataset is appropriate for personalized saliency detection task, we compare the inter-subject consistency, *i.e.*, the agreement among different viewers, in our PSM dataset and other related datasets. Specifically, for each dataset, we first enumerate all possible subject-pairs, *i.e.*, two different subjects, and then compute the average CC, Similarity, AUC-Judd, and NSS scores across all pairs. Recall that our PSM dataset consists of images from different datasets, *e.g.*, MIT, OSIE, ImageNet, PASCAL-S, SALICON, iSUN *etc.*, and only MIT, OSIE, PASCAL-S are designed for saliency tasks[2]. Hence, we only compare the consistency scores among ours and above three datasets and show the results in Table 5. We observe that our dataset achieves the lowest inter-subject consistency values among all relative ones, indicating that the heterogeneity in our saliency maps is more severe than that in the others.

We compute the within-subject consistency and the inter-subject consistency as shown in Table 4. Within-subject consistency measures the agreement of saliency maps viewed at different time by the same observer. We can see that the within-subject consistency is high and the inter-subject consistency is low in our dataset, which hints the eye fixations are subject-dependent.

#### 3.3.3 Ours vs. Existing Datasets

Recall that our objective is to identify the heterogeneity of fixation maps across individuals. Therefore it is crucial that we obtain reliable ground truth annotation. Existing datasets, such as MIT, OSIE, do provide individual eye fixation information, but each image is viewed only once. In Section 3.3 and Figure 3, we empirically show that viewing each image multiple times leads to a more reliable ground truth. Thus in our dataset, each image is viewed 4 times with at least a one-day time interval. Furthermore, as aforementioned, since fine-tuning a separate network for each participant is not scalable in real applications, we leverage person-specific information for personalized saliency prediction. However, such person-specific information is not available in all existing datasets.

---

2. Even though SALICON and iSUN are also eye fixation datasets, the ground truth is annotated based on mouse-tracking and web camera, respectively.

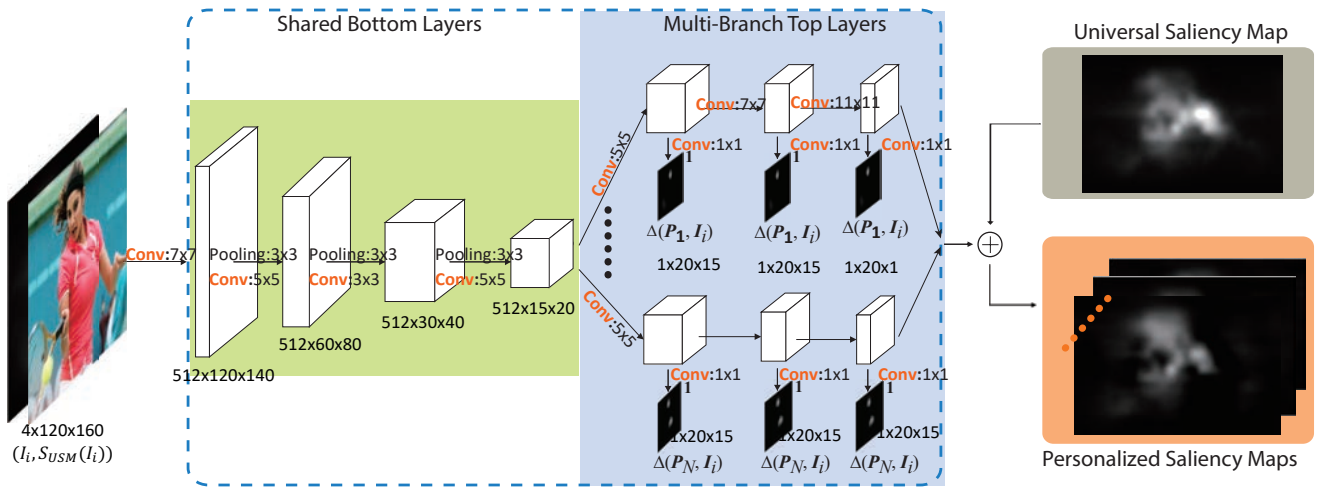

Fig. 4: The pipeline of our Multi-task CNN based PSM prediction. The input is the image with its predicted USM. Specifically, we treat the discrepancy prediction for each person as a separate task. There are $n$ persons in our dataset, then there are $n$ tasks in this framework. We then sum the predicted discrepancy map with USM and generate the final estimated PSM.

# 4 OUR APPROACH

## 4.1 USM based PSM Formulation

Many existing approaches [19] [10] have employed CNN in an end-to-end learning strategy to predict learning saliency map and achieve state-of-the-art performance. Intuitively, we can follow the same strategy for PSM prediction, *i.e.*, training a separate CNN for each participant to map the RGB images to PSMs. However, such strategy is neither scalable nor feasible for a number of reasons. Firstly, it needs a vast amount of training samples to learn a robust CNN for each participant. This requires each participant to view thousands of images with high concentration, which is hard and extremely time consuming. Secondly, training multiple CNNs for different participants is computationally expensive and inefficient.

While each participant is unique in terms of gender, age, personal preference, *etc.*, resulting in incongruity in saliency preference, different participants still share commonalities in their observed saliency maps because certain objects, such as faces and logos, always seem to attract the attention of all participants as shown in Fig. 1.

For this reason, instead of directly predicting the PSM, we set out to explore the difference map between USM and PSM. The discrepancy map $\Delta(P_n, I_i)$ for the given image $I_i$ ($i = 1, \ldots, K$) and the $n$-th participant $P_n$ ($n = 1, \ldots, N$) is of the form:

$$S_{PSM}(P_n, I_i) = S_{USM}(I_i) + \Delta(P_n, I_i) \qquad (1)$$

where $S_{PSM}(P_n, I_i)$ is the desired personalized saliency map and $S_{USM}(I_i)$ is the universal saliency map.

Note that USMs predicted by traditional saliency method entail the commonality observed by different participants. We convert the problem of predicting PSMs to estimating the discrepancy $\Delta(P_n, I_i)$ because that universal saliency map $S_{USM}(I_i)$ itself already provides a rough estimation of the PSM whereas the discrepancy $\Delta(P_n, I_i)$ would serve as an error correction function. Previous work has shown that error correction scheme works well for CNN based regression task [31] and classification task [40], which motivates us to model PSM based on USM.

## 4.2 Discrepancy Prediction

We observe, on one hand, such $\Delta(P_n, I_i)$ is closely related to the contents of the input image; On the other, $\Delta(P_n, I_i)$ is subject-dependent. Previous work observes that human attention is related to several observer-dependent factors. For example, personal preference [32] [33], gender [34], age [35] are all closely related to each individual's visual attention in a given image. Specifically, there is a study showing that female observers would pay more attention to clothes and shoes than male [36]. Therefore, in this paper, we propose two solutions to predict such $\Delta(P_n, I_i)$: i) when we do not have access to the observer's personal information, we leverage a Multi-task CNN scheme to learn a separate model for discrepancy prediction for each observer; ii) when we do have the observer's personal information, we encode his/her personal information with CNN filters, and propose a CNN with Person-specific Information Encoded Filters (CNN-PIEF) scheme for discrepancy prediction. The two approaches have their own cons and pros as we will discuss later.

It is worth noting that to predict the discrepancy, we take the predicted USM as the input of the network. There are two reasons for this. Firstly, the predicted USM provides a reference of the discrepancy. In other words, the discrepancy is a function of predicted USM. The discrepancy prediction without USM would be difficult. Secondly, the predicted USM is a coarse estimation of PSM, and the strategy of using it for discrepancy prediction is inspired by the iterative error feedback in pose estimation [56], instance segment [57], which have shown the effectiveness of concatenating the input image and coarse prediction to predict the discrepancy for both pose points estimation and segmentation.

### 4.2.1 Multi-task CNN based Discrepancy Prediction

Previous approaches [21] [22] have shown that features extracted by the first several layers can be shared by related tasks. Actually, the discrepancy prediction for the given image is distinct but related regression tasks across different observers, which motivates our multi-task CNN solution. In our Multi-task CNN based discrepancy prediction, the discrepancy prediction for different observers corresponds to different tasks. The inputs of Multi-task CNN network are images with their corresponding universal saliency maps and our goal is to estimate the discrepancy maps $\Delta(P_n, I_i)$, $i = 1, \ldots, k$ for the $n$-th participant through the $n$-th task. The network architecture of our Multi-task CNN is illustrated in Fig. 4.

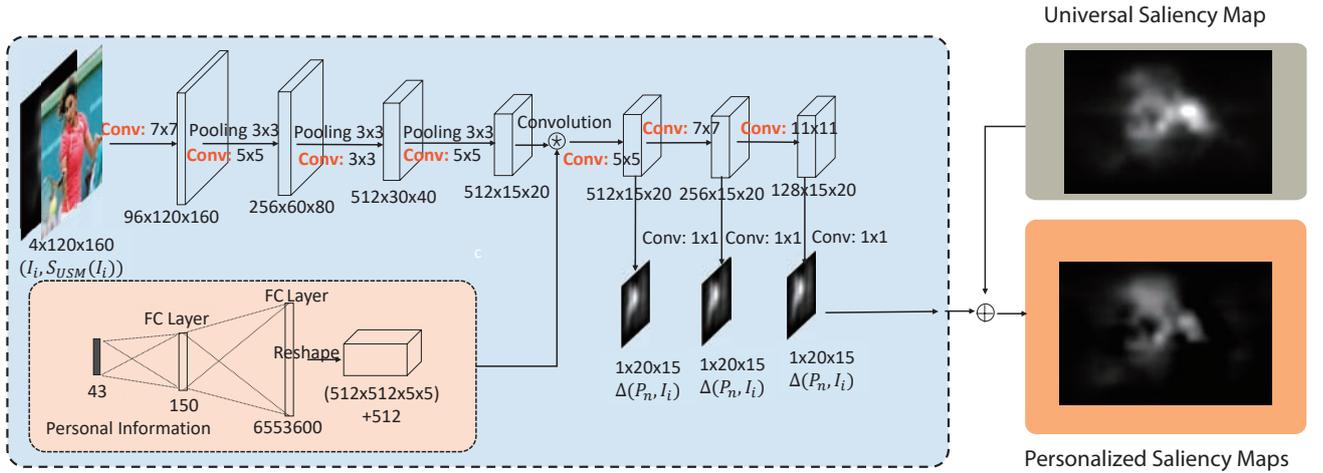

Fig. 5: The pipeline of our PSM prediction model. The input is the image with its predicted USM. Specifically, we embed the person-specific information into CNN by encoding it into filters and then convolving filters with the output of *conv4* layer. We then sum the predicted discrepancy map with USM and get the final estimated PSM.

Suppose we have $N$ participants in total. We concatenate an RGB image of $160 \times 120$ pixels with its USM predicted by general saliency model and get a $160 \times 120 \times 4$ cube and use it as the input of the Multi-task network. For image $I_i$, $\Delta(P_n, I_i)$ is the output corresponding to the discrepancy between PSM and USM for the $n$-th person. There are four convolutional layers shared by all participants after which the network is then split into $N$ tasks which are exclusive for $N$ participants. Each task has three convolutional layers followed by a ReLU activation function.

[19] and [20] show that by adding the supervision in the middle layers, the features learned by CNN will be more discriminative, and might boost the performance of a given task. Consequently, we add an additional loss layer after the *conv5* and *conv6*, respectively, for each task, which can help the prediction of $\Delta(P_n, I_i)$. For the $n$-th task, $f_\ell^n(S_{USM}(I_i), I_i) \in \mathbb{R}^{h_\ell \times w_\ell \times d_\ell}(\ell = 5, 6, 7)$ is the feature map after the $\ell$-th convolutional layer (the first convolutional layer corresponds to the first exclusive convolutional layer, so $\ell$ starts from 5). For each feature map $f_\ell^n(S_{USM}(I_i), I_i)$, a $1 \times 1$ convolutional layer is employed to map it to $S_\ell(S_{USM}(I_i), I_i) \in \mathbb{R}^{h_\ell \times w_\ell \times 1}$, which is the target discrepancy. To make $S_\ell(S_{USM}(I_i), I_i)$ close to $\Delta_\ell(P_n, I_i)$, we set the objective function as:

$$\min \quad \sum_{\ell=5}^{7} \sum_{n=1}^{N} \sum_{i=1}^{K} \|S_k(S_{USM}(I_i), I_i) - \Delta_\ell(P_n, I_i)\|_F^2 \quad (2)$$

Then we use a mini-batch based stochastic gradient descent to optimize all parameters in our Multi-task CNN.

It is important to note that multi-task CNN based discrepancy prediction learns separate but related models for different observers. Compared with CNN-PIEF, multi-task CNN cannot exploit which personal information is related to PSM prediction. In the deployment stage, given any unrecorded observer, he/she has to view lots of images to get the PSMs. Then we use these images and their PSMs to retrain the network with an additional task, which is time-consuming.

### 4.2.2 CNN-PIEF based Discrepancy Prediction

Since $\Delta(P_n, I_i)$ is observer-dependent and also relies on the contents of the input image, we propose to modify the network from [31] and use person-specific information to generate filters and convolve these Person-specific Information Encoded Filters (PIEF) with the input feature maps, and these PIEF function as "switches" to determine which areas are interested by different observers. The inputs of the network are images with their corresponding universal saliency maps and the observer-dependent PIEF, the goal is to estimate the difference/discrepancy maps $\Delta(P_n, I_i)$ for each participant.

For each individual, we conduct a simple survey to acquire his/her personal information and then encode the collected information by one-hot code. The information of different observers then is coded with vectors with the same length. By following the recent work of [37] [38] [39], we propose to embed our person-specific information into CNN, *i.e.*, we reshape the features in the last layer into a set of convolutional filters (4D tensor) which is used to convolve with the output from the *conv4* layer in our CNN.

Suppose that we have $N$ participants in total, for image $I_i$, the output of CNN-PIEF corresponds to the discrepancy between PSM and USM for the $n$-th person: $\Delta(P, I_i)$. We concatenate a RGB image of $160 \times 120$ with its USM predicted by some existing saliency model, resulting in a $160 \times 120 \times 4$ cube, as the input of CNN. We use $f_\ell(S_{USM}(I_i), I_i) \in \mathbb{R}^{h_\ell \times w_\ell \times d_\ell}(\ell = 5, 6, 7)$ to donate the feature map after the $\ell$-th convolutional layer. Inspired by previous work [19], [20], middle layer supervision is imposed by adding additional loss layer after *conv5* and *conv6* layers, respectively. For each feature map $f_\ell(S_{USM}(I_i), I_i)$, we use a $1 \times 1$ convolutional layer to map it to a feature map $S_\ell(S_{USM}(I_i), I_i) \in \mathbb{R}^{h_\ell \times w_\ell \times 1}$ which corresponds to the predicted discrepancy. It is desirable that $S_\ell(S_{USM}(I_i), I_i)$ is close to $\Delta_\ell(P, I_i)$ which is obtained by resizing to $\Delta_\ell(P_n, I_i)$ to the size of $h_\ell \times w_\ell \times 1$. Thus we arrive at the following objective function:

$$\min \quad \sum_{\ell=5}^{7} \sum_{i=1}^{K} \|S_k(S_{USM}(I_i), I_i) - \Delta_\ell(P, I_i)\|_F^2 \quad (3)$$

The network architecture of our CNN-PIEF is illustrated in Fig. 5.

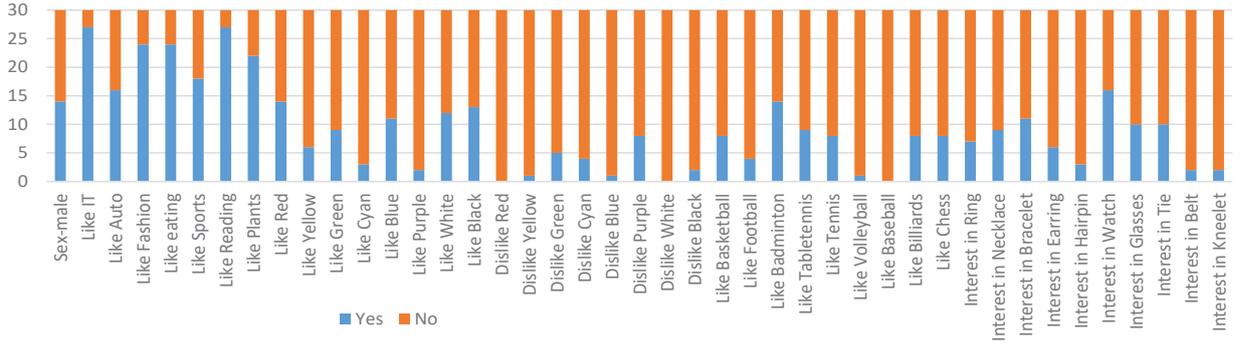

Fig. 6: The statistics of our survey results (best viewed in color)

**Person-specific Information Collection**

Recall that PSM prediction task is challenging because too many factors contribute to the saliency variations across different individuals, and the investigation of key factors leading to the observer-dependent saliency falls into the regime of psychology. In this paper, we don't aim to find out all factors that facilitate the attention discrepancy. Instead, the goal of our work is to show that by encoding a fraction of personal information of participants into CNN, we can approximately model the visual attention pattern of each individual (personalized saliency).

The personal information is comprised of the following parts: Firstly, previous studies have shown that gender [36] causes the heterogeneity of saliency map across different observers [3], which can also be observed in our experiments. Therefore, we collect the gender information of all observers. Secondly, considering that color is an important factor in universal saliency detection, and our conjecture that different preference/sensitive of different observers to different colors may cause the inconsistence of saliency, we also collect the preference of different observers over limited numbers of colors. Specifically, we choose three primary colors (*i.e.*, Red, Green, and Blue), additive secondary colors (*i.e.*, Cyan, Magenta, and Yellow), Black and White, and collect the preference/disgust of all subjects over these colors. Thirdly, we conjecture that the educational background and the hobbies of different observers may also result in their different interestingness to different objects. Therefore, we rank all the objects by their observer-dependency attention variance on the training set and pick the top ranked ones to generate PIEF.

In order to calculate the distribution of the interestingness of various objects for different participants, we manually annotate all the objects (242 classes in total) in our dataset. Mathematically, for the $n$-th observer $P_n$ ($n = 1, \ldots, N$), given an image $I_i$ ($i = 1, \ldots, K$), we donate the personalized saliency maps (PSM) of the $n$-th participant to $I_i$ as $S_{PSM}(P_n, I_i)$, and denote the binary mask of a given class $C_j$ ($j = 1, \ldots, 242$) in image $I_i$ as $M_{C_j, I_i}$, we use $\mathbf{Int}(P_n, I_i, C_j)$ to represent the interestingness of class $C_j$ to participant $P_n$ in image $I_i$, which is calculated as:

$$\mathbf{Int}(P_n, I_i, C_{j, I_i}) = \frac{\|\mathbf{vec}(S_{PSM}(P_n, I_i) \odot M_{C_j, I_i})\|_1}{\|\mathbf{vec}(S_{PSM}(P_n, I_i))\|_1} \quad (4)$$

3. Psychology studies [36] also show that age is the factor causing the heterogeneity of saliency maps across different observers, as the ages of participants in our study fall in a small range of 20-25 years old, we don't include it for PIEF.

We then define the interestingness of class $C_j$ to observer $P_n$ as $\mathbf{Int}(P_n, C_j)$, which can be calculated as follows:

$$\mathbf{Int}(P_n, C_j) = \frac{1}{Z} \sum_{I_i} \mathbf{Int}(P_n, I_i, C_j) \quad (5)$$

where $Z$ is the number of images that contain the annotated objects in class $C_j$. $\mathbf{Int}(P_n, C_j)$ measures the interestingness of the segment from class $C_j$ to the observer $P_n$. We rank all the objects by their observer-dependency attention variance in the training set. The objects with large attention variance can be roughly categorized as fashions, sports, texts, *etc.*, as shown in Fig. 2.

Then we conduct a survey to collect each observer's personal information. Specifically, we collect observer's gender information (1D), the preference to objects falling into the fashion category (ring, necklace, bracelet, earring, hairpin, watch, glasses, tie, belt, kneelet *etc.*, 11D), the preference/disgust to colors (red, yellow, green, cyan, blue, purple, white, black, 16D), the preference to different sports (football, basketball, badminton, tabletennis, tennis, volleyball, baseball, and billiards *etc.*, 11D), and the preference to objects falling into other categories (IT, plant, texts, food, 4D). One-hot encoding strategy is used to encode each observer's personal information. The dimensionality of encoded information vector for each observer is 43D. Then we feed such personal-information into our CNN-PIEF for PSM prediction.

The personal information survey is conducted after eye tracking. In order to reduce the carry-over effect as much as possible, we enforce each session on a different day of the week and the time interval for the same person to view the same image ranging from one day to one week. In addition, images in each session are randomly shown on the screen to decrease the impact of long-term memory of the person for the given image in a fixed sequence. Fig. 6 shows the statistics of our survey results. We ask each subject to answer a questionnaire that includes the following two types of questions: 1) yes-no questions: "Do you like IT?" "Do you like cars?", "Do you like fashion?", "Do you enjoy eating?", "Do you like sports?", "Do you like reading?" and "Do you like plant?"; 2) questions with (possible) multiple choices: "What kind of colors do you like?", "What kind of colors do you disgust?", "What kind of sports do you like?", "Which kind of jewelries/accessories do you like?". For personal preferences with respect to colors, since there are two questions (like or disgust) about colors, each color actually can be rated as "like", "disgust", or "neither like nor disgust" by each subject. In addition, regarding the question "Which color do you like?", each subject can select multiple colors rather than only one where the choices are provided with both color words (like 'red', 'green', 'blue') and color stimuli (patches with the color).

**Additional remarks:** On one hand, the performance of CNN-PIEF depends on the choice of person-specific information, while factors leading to personalized saliency are still not clear in psychology, which restricts the performance of our model. However, our experiments show even with a small fraction of personal information for PIEF, our model still achieves comparable performance compared with Multi-task CNN based discrepancy prediction, as discussed in Section 5.2; On the other hand, in our CNN-PIEF, all individuals share the same network. Therefore, in the deployment stage, given any unrecorded observer, our model requires only for his/her person-specific information to for PSM prediction. The high scalability property of our network makes it easy to be deployed our network in real applications, as discussed in Section 5.3.

# 5 EXPERIMENTAL VALIDATIONS

## 5.1 Experimental Setup

**Parameters.** We implement our solutions with the CAFFE framework [23]. To avoid over-fitting and improve model robustness, we augment the training data through left-right flip operations.

In multi-task CNN network, we train our network with the following hyper-parameters: mini-batch size (40), learning rate (0.0003), momentum (0.9), weight decay (0.0005), and the number of iterations (40,000). The parameters corresponding to the universal saliency map channel and $1 \times 1$ *conv* layers for middle layer supervision are initialized with 'xavier'. Following the same initialization step in [10] and [13], we use the well-trained DeepNet model to initialize the corresponding parameters in our network. The network architecture of our Multi-task CNN is identical to that of DeepNet [10] except that: i) the parameters corresponding to tasks of different observers are different; ii) middle layer supervision is imposed by adding $1 \times 1$ *conv* layer after *conv5* and *conv6*, respectively; iii) an additional channel corresponding to USM is added in the input.

In CNN-PIEF, we train our network with the following hyper-parameters: mini-batch size (40), learning rate ($1e$-6), momentum (0.9), weight decay (0.0005), number of iterations (100,000). The parameters are initialized with 'xavier'.

**Evaluation Protocols.** We evaluate our approach under two settings: **closed-set** setting and **open-set** setting. Specifically, we randomly choose 20 observers whose PSMs are used for closed-set evaluation, while the PSMs corresponding to the remaining 10 observers are used for open-set evaluation. Of all 1600 images, we randomly choose 1000 images and use them as training images, and use the remaining 600 images as testing images. *The closed-set setting is used to valuate how well our model can predict the visual attention pattern of seen observers.* Therefore we use the 1000 images corresponding to those 20 observers to train a model, and evaluate the model with the remaining 600 images associated with these 20 observers. *The open-set setting is used to evaluate whether our model is transferable,* i.e., *predicting the fixation maps for an unseen observer.* In the open-set setting, we evaluate the model trained in the closed-set setting with the 600 images corresponding to those 10 unseen objects in the open-set.

**Measurements.** By following [27] [10] [13], we choose CC, Similarity, AUC-Judd, and NSS [18] to measure the differences between the predicted PSM and its corresponding ground truth.

**Baselines.** In our model, all existing fixation prediction methods can be used to generate USMs, we then concatenate predicted UMS with RGB image as the input of our network. Based on the performance of existing methods on the MIT saliency benchmark [24] in terms of similarity, we choose LDS [25], BMS [26], ML-Net [19], and SalNet [10] to predict the USMs for images in our dataset. The first two methods are based on hand-crafted features, and the latter two are based on deep learning techniques. We use the codes provided by the authors of these methods to generate USM.

To validate the effectiveness of our models, we compare our Multi-CNN and CNN-PIEF with the following baseline algorithms:

- **RGB based MultiConvNets**: Different ConvNets are trained to predict $\Delta(P_n, I_i)$ for different observers independently, with RGB images as inputs.
- **RGB based Multi-task**: Multi-task CNN architecture is trained to predict $\Delta(P_n, I_i)$ for all participants simultaneously, with RGB images as inputs.
- $X$ **based MultiConvNets**: Different ConvNets are trained to predict $\Delta(P_n, I_i)$ for different observers independently, with RGB images and USM predicted by method $X$ as inputs, where $X$ donates LDS, BMS, ML-Net, and SalNet, respectively.

In order to show the upper limit of our method, we train the following networks by taking the Ground Truth USM (GT USM) as input for discrepancy prediction:

- **GT USM based MultiConvNets**: ConvNets are trained to predict $\Delta(P_n, I_i)$ for different observers independently, whose inputs are RGB images and GT USM.
- **GT USM based CNN-PIEF**: CNN-PIEF is trained to predict $\Delta(P_n, I_i)$ for different observers independently, whose inputs are RGB images, personal information and GT USM.
- **GT USM based Multi-task CNN**: Multi-task CNN is trained to predict $\Delta(P_n, I_i)$ for different observer independently, whose inputs are RGB images and GT USM.

We have trained a universal saliency model (SalNet) with our dataset for USM prediction for different methods, then we use the USM predicted by the model trained on our dataset for discrepancy prediction. We denote such baselines as **SalNet (FT) based MultiConvNets**, **SalNet (FT) based Multi-task CNN**, and **SalNet (FT) based CNN-PIEF**, respectively.

We further compare our methods with the following baselines that jointly predict the USM and the discrepancy maps under closed-set setting:

- **Baseline 1**: ConvNets are trained to jointly predict both USM and $\Delta(P_n, I_i)$ for different observers independently, with RGB images as inputs.
- **Baseline 2**: CNN-PIEF is trained to jointly predict both USM and $\Delta(P_n, I_i)$ for different observers independently, with RGB images and personal information as inputs.
- **Baseline 3**: Multi-task CNN is trained to jointly predict both USM and $\Delta(P_n, I_i)$ for different observers independently, with RGB images as inputs.

It is worth noting that the network architectures of these baselines are similar to ours. The only differences are the number of input channels and whether the parameters are shared in the

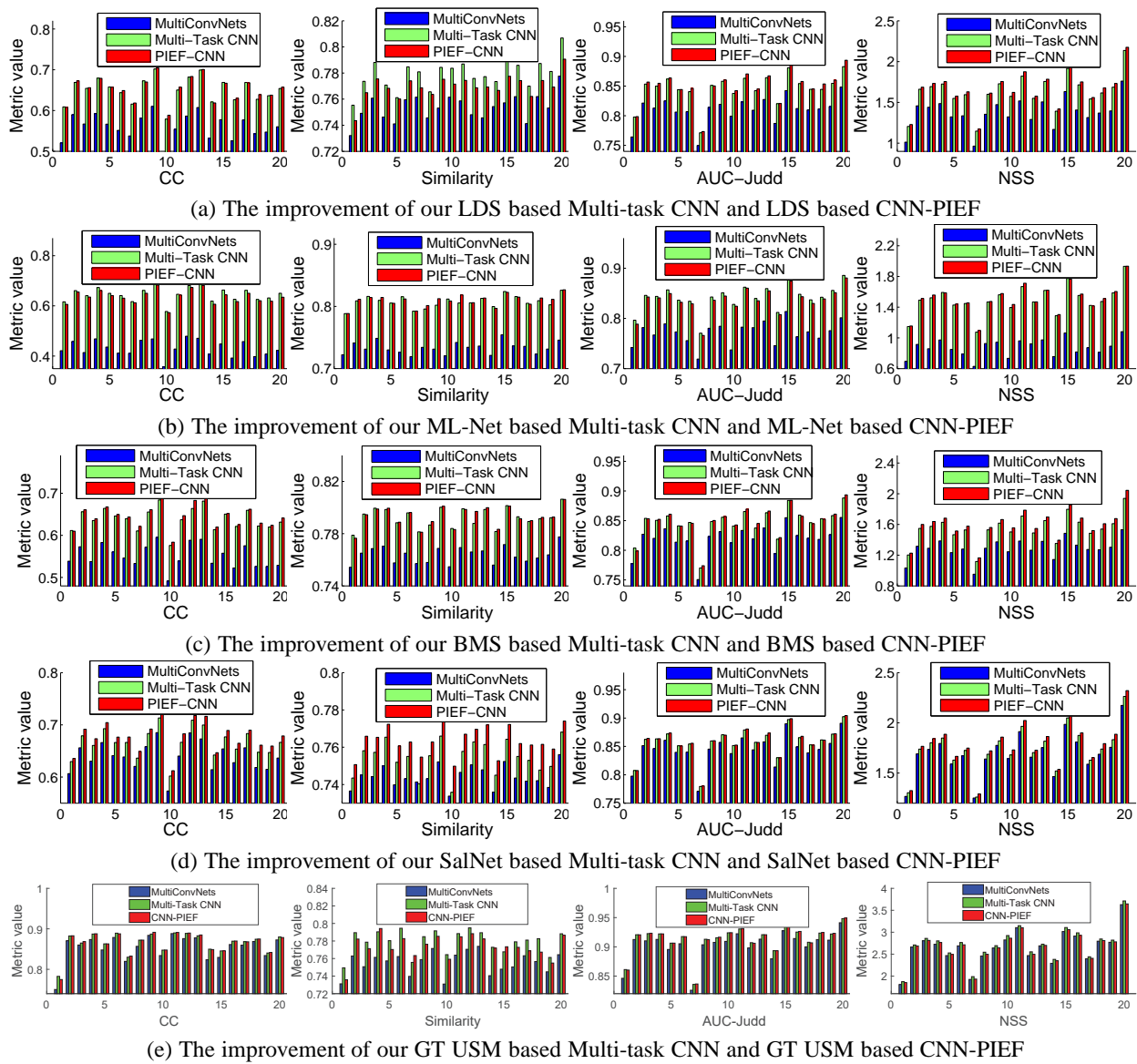

(a) The improvement of our LDS based Multi-task CNN and LDS based CNN-PIEF

(b) The improvement of our ML-Net based Multi-task CNN and ML-Net based CNN-PIEF

(c) The improvement of our BMS based Multi-task CNN and BMS based CNN-PIEF

(d) The improvement of our SalNet based Multi-task CNN and SalNet based CNN-PIEF

(e) The improvement of our GT USM based Multi-task CNN and GT USM based CNN-PIEF

Fig. 7: MultiConvNets vs. our methods for PSM predictions for each observer. In each graph, $x$ axis represents the ID of each observer and $y$ axis represents the metric value of CC, Similarity, AUC-Judd or NSS. We can see that our methods always outperform MultiConvNets under closed-set setting.

first few layers. For fair comparisons, we have employed the same strategies in terms of data augmentation, middle layers supervision, and parameter initializations for all these baselines.

### 5.2 Performance Evaluation Under Closed-Set Setting

The performance of all methods under closed-set setting are listed in Table 6. Our solutions always outperform the baseline methods in terms of all metrics, which demonstrates the effectiveness of our method. Furthermore, the discrepancy based personalized saliency detection methods consistently outperform directly predicting PSM from RGB images. This validates the effectiveness of our "error correction" strategy for personalized saliency detection.

In addition, by comparing the performance of our method with baseline 1, 2, 3 that jointly predict the USM and the discrepancy maps under the closed-set setting, we can see that the predicted discrepancy maps with jointly training methods do not bring additional performance improvement over USM. The reason is that the discrepancy is related to the predicted USM, *i.e.*, it is a function of predicted USM. Without predicted USM as input, it is difficult to predict discrepancy because the reference of the discrepancy is unknown.

Table 6 shows that GT USM based models correspond to better performance, which means USM prediction is extremely important for PSM prediction. With better USM, our PSM can be further improved. Further, Table 6 shows that fine-tuning SalNet with our dataset would lead to worse performance for both closed-set setting and open-set setting. The possible reason is that original SalNet is pre-trained on SALICON with 10,000 images, so the generalization capability is good. If it is fine-tuned with our dataset, which only contains 1000 images, the fine-tuned USM prediction model would overfit these 1000 images, consequently the generalization capability of the model would be reduced. Without a good USM, PSM prediction would drop too.

| Methods | CC | Similarity | AUC-Judd | NSS |
|---|---|---|---|---|
| RGB based MultiConvNets | 0.5832 | 0.6425 | 0.8382 | 1.3689 |
| RGB based CNN-PIEF | 0.6109 | 0.6513 | 0.8483 | 1.4763 |
| RGB based Multi-task CNN | 0.6013 | 0.7172 | 0.8465 | 1.4771 |
| USM predicted by Baseline 1 | 0.5892 | 0.6666 | 0.8380 | 1.3866 |
| PSM predicted by Baseline 1 | 0.5703 | 0.6431 | 0.8390 | 1.6127 |
| USM predicted by Baseline 2 | 0.6335 | 0.6966 | 0.8470 | 1.3866 |
| PSM predicted by Baseline 2 | **0.6399** | **0.6986** | **0.8508** | **1.6325** |
| USM predicted by Baseline 3 | 0.6065 | 0.7034 | 0.8443 | 1.5955 |
| PSM predicted by Baseline 3 | **0.6033** | **0.6952** | **0.8479** | **1.6400** |
| LDS [25] | 0.5481 | 0.5777 | 0.8264 | 1.3274 |
| LDS based MultiConvNets | 0.5610 | 0.7535 | 0.8103 | 1.3767 |
| LDS based CNN-PIEF | **0.6532** | 0.7696 | **0.8494** | **1.6638** |
| LDS based Multi-task CNN | 0.6509 | **0.7792** | 0.8459 | 1.6308 |
| ML-Net [19] | 0.3198 | 0.4941 | 0.7117 | 0.5664 |
| ML-Net based MultiConvNets | 0.4310 | 0.7333 | 0.7711 | 0.8733 |
| ML-Net based CNN-PIEF | 0.6368 | **0.8095** | 0.8365 | **1.5105** |
| ML-Net based Multi-task CNN | **0.6463** | 0.8077 | **0.8414** | 1.4960 |
| BMS [26] | 0.4937 | 0.6757 | 0.8009 | 1.1241 |
| BMS based MultiConvNets | 0.5510 | 0.7636 | 0.8196 | 1.2884 |
| BMS based CNN-PIEF | **0.6448** | **0.7931** | **0.8486** | **1.6004** |
| BMS based Multi-task CNN | 0.6390 | 0.7925 | 0.8472 | 1.5464 |
| SalNet [10] | 0.6238 | 0.6847 | 0.8471 | 1.5848 |
| SalNet based MultiConvNets | 0.6397 | 0.7442 | 0.8448 | 1.6924 |
| SalNet based CNN-PIEF | **0.6771** | **0.7636** | **0.8588** | **1.7819** |
| SalNet based Multi-task CNN | 0.6661 | 0.7547 | 0.8580 | 1.7445 |
| SalNet(FT) | 0.5060 | 0.5924 | 0.8041 | 1.2218 |
| SalNet(FT) based MultiConvNets | 0.5516 | 0.7241 | 0.8109 | 1.3207 |
| SalNet(FT) based CNN-PIEF | **0.6288** | **0.7481** | **0.8411** | **1.5702** |
| SalNet(FT) based Multi-task CNN | 0.5745 | 0.7313 | 0.8217 | 1.3637 |
| GT USM | 0.8563 | 0.7189 | 0.9103 | 2.6181 |
| GT USM based MultiConvNets | 0.8543 | 0.7571 | 0.9031 | 2.6581 |
| GT USM based CNN-PIEF | **0.8651** | 0.7736 | **0.9129** | 2.6775 |
| GT USM based Multi-task CNN | 0.8648 | **0.7792** | 0.9103 | **2.7207** |

TABLE 6: The performance comparison of different methods under closed-set settings on our PSM dataset. (FT) means the models are fine-tuned with our training set.

| Methods | CC | Similarity | AUC-Judd | NSS |
|---|---|---|---|---|
| LDS [25] | 0.5765 | 0.5839 | 0.8528 | 1.5336 |
| LDS based CNN-PIEF | 0.6638 | 0.7771 | 0.8693 | 1.8147 |
| LDS based CNN-PIEF (transferred model) | **0.6284** | **0.7626** | **0.8601** | **1.7239** |
| ML-Net [19] | 0.3317 | 0.4973 | 0.7350 | 0.6636 |
| ML-Net based CNN-PIEF | 0.6450 | 0.8166 | 0.8559 | 1.6879 |
| ML-Net based CNN-PIEF (transferred model) | **0.6117** | **0.7946** | **0.8534** | **1.5490** |
| BMS [26] | 0.4637 | 0.6641 | 0.8008 | 1.1531 |
| BMS based CNN-PIEF | 0.6506 | 0.7995 | 0.8685 | 1.7864 |
| BMS based CNN-PIEF (transferred model) | **0.6038** | **0.7818** | **0.8629** | **1.5978** |
| SalNet [10] | 0.5981 | 0.6699 | 0.8629 | 1.6701 |
| SalNet based CNN-PIEF | 0.6863 | 0.7688 | 0.8811 | 1.9877 |
| SalNet based CNN-PIEF (transferred model) | **0.6468** | **0.7519** | **0.8740** | **1.8484** |
| SalNet (FT) | 0.5054 | 0.5890 | 0.8250 | 1.3388 |
| SalNet (FT) based CNN-PIEF | 0.6541 | 0.7577 | 0.8684 | 1.7903 |
| SalNet (FT) based CNN-PIEF (transferred model) | **0.5801** | **0.7282** | **0.8462** | **1.5927** |
| GT USM | 0.8516 | 0.6866 | 0.9274 | 2.8076 |
| GT USM based CNN-PIEF | 0.8681 | 0.7771 | 0.9306 | 2.9100 |
| GT USM based CNN-PIEF (transferred model) | **0.8431** | **0.7524** | **0.9264** | **2.8349** |

TABLE 7: The performance comparison of different methods under open-set setting on our PSM dataset. The baselines in this table are trained with 1000 images corresponding to the 10 participants under the open-set setting, while the models are trained with the training data under our closed-set setting

on USM prediction. If ground truth USM is used for discrepancy prediction, the performance would be greatly improved as shown in Table 6, which means with better USM prediction methods, our PSM can be further improved.

### 5.3 Performance Evaluation Under Open-Set Setting

To evaluate the generalization capability of our model, we directly adopt the model trained under closed-set setting and test it with the 600 testing images corresponding to the 10 unseen observers in the open-set. The performance of different methods is shown in Table 7. The baselines ($X$ based CNN-PIEF) in Table 7 are trained with the 1000 training images corresponding to the 10 observers under the open-set setting, while the "transferred model" corresponds to the model trained on 20 observers under the closed-set setting.

Despite being trained with PSMs on totally different observers, the transferred models could beat USM in terms of all metrics, while still maintains a comparable performance against models that are specifically trained for the 10 observers. It's worth mentioning that the transferred model is only trained with a small dataset of 20 observers, further performance improvement can be expected with a larger dataset with more observers. The good performance of the transferred models in this experiment demonstrates the scalability of our model for PSM prediction, *i.e.*, once we collect the PSMs and their person-specific information for a large amount of observers, a model with good generalization capability can be trained. For a new observer, we can easily collect his/her person-specific information, enabling us to predict the PSM of him/her over any given image. It takes about 10 hours for each participant to observe 1K testing images 4 times. On the contrary, person-specific information can be easily acquired by answering a few questions, which can be done in only a few minutes. Further, even we collect some training images for a new observer, because of the limited training samples, it may not contain sufficient personal information for personalized saliency prediction.

### 5.4 The effect of supervision on middle layers

Fig. 9 shows the accuracy gain in CC, Similarity, AUC-Judd and NSS metrics from imposing supervision on middle layers in our

We further show the performance of different PSM prediction methods for each observer on the testing set in Fig. 7. Both Table 6 and Fig. 7 show that our Multi-task CNN scheme and CNN-PIEF scheme always show higher performance for fixation prediction than simply training a CNN for each observer. Compared with MultiConvNets strategy, our Multi-task CNN and CNN-PIEF have the following advantages: i) in both CNN-PIEF and Multi-task CNN, some/all network parameters are actually shared across different observers. Since the PSM prediction for different observers are related tasks, and our solutions greatly reduce the number of parameters to be learned. Given the limited number of training samples, our solutions help train a more robust network for both Multi-CNN and CNN-PIEF, consequently boosts the PSM prediction preformation, which agrees with existing work for multi-task learning [21] [22]. However, given the limited number of training samples, it is unlikely to train a robust network for each observer in MultiConvNets; ii) it is also worth noting that such MultiConvNets based approach is also not scalable for PSMs prediction in real applications where there are many observers, because both the collection of training data and the phase of training multiple networks is time consuming. In contrast, our solution can reduce the time costs in network training. Further, as shown in the following section (Section 5.3), for an unseen observer, our CNN-PIEF model can be easily transferred to this new observer and achieve satisfactory performance, which further validates the scalability of our solutions.

We also show some predicted saliency maps for different participants in Fig. 8. It is worth noting that though our predicted PSMs are a bit noisy and blurry, our work is the first attempt along this direction. As aforementioned, our method greatly relies

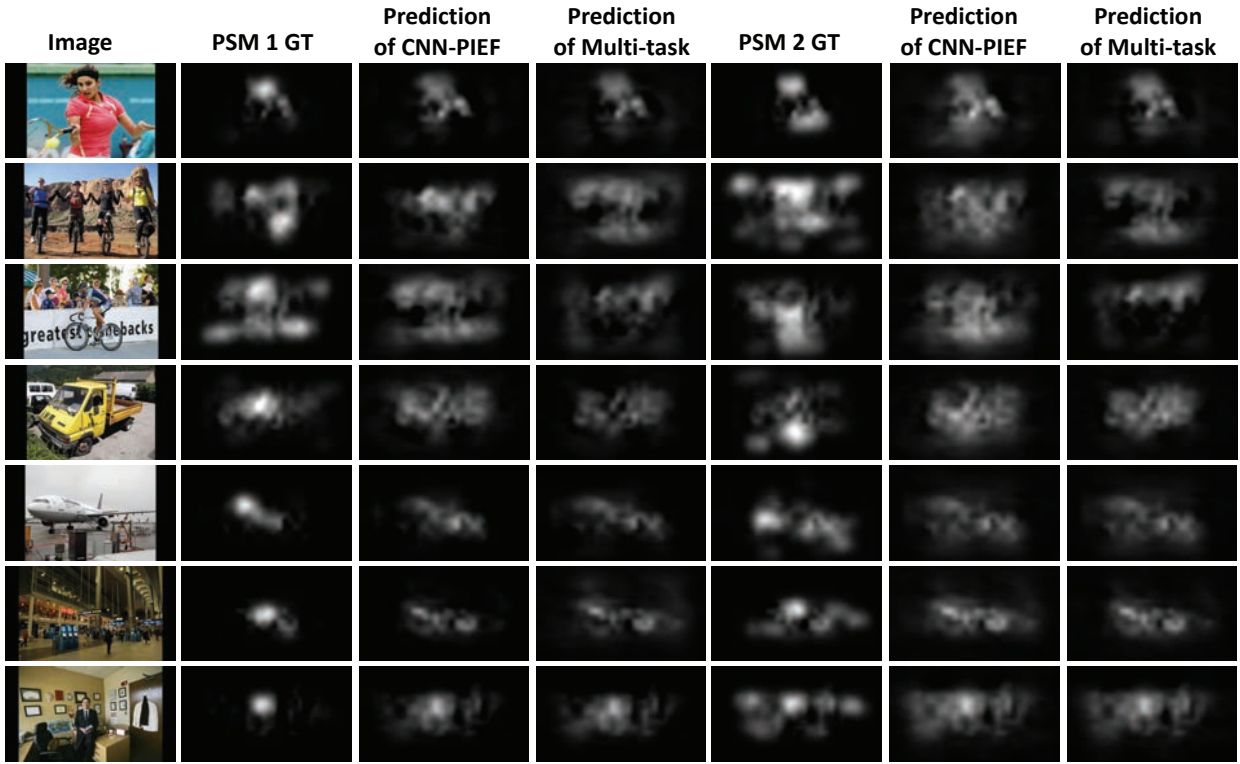

Fig. 8: Some examples of PSMs predicted by our methods.

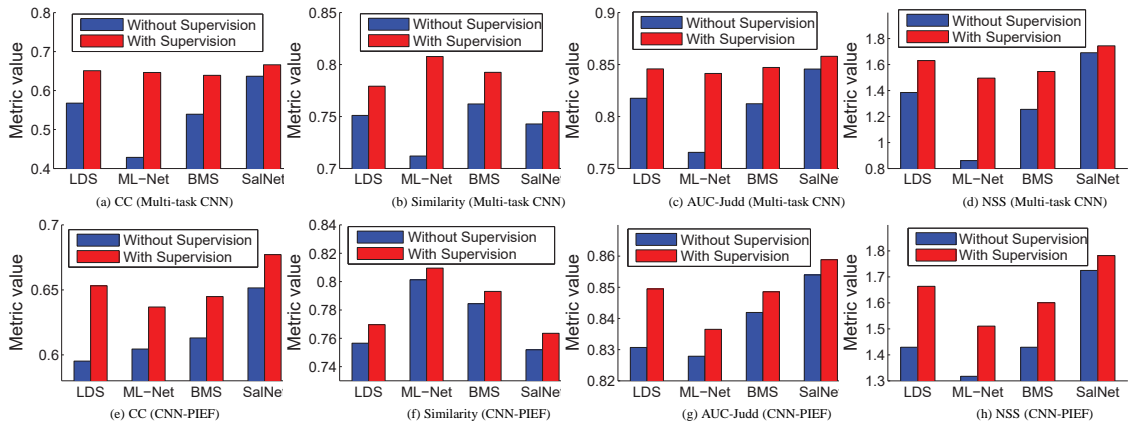

Fig. 9: The effect of supervision on middle layers in our Multi-task CNN and CNN-PIEF.

Multi-task CNN and CNN-PIEF based on LDS, ML-Net, BMS, and SalNet. We observe that middle layer supervision is helpful for PSM prediction in line with previous findings in [20]. By comparing Fig. 9 with Table 6, we can see that even without supervision on middle layers, our solutions still outperform all baselines, which further validates the effectiveness of our models.

### 5.5 The Position of Inserting Person-Specific Information

We further empirically investigate how the position of the person-specific information encoded filters would affect the performance of our network. As shown in Table 9, inserting filters in *conv5* layer yields the best performance. We also observe a descending trend in terms of all measurements by inserting the person-specific information in latter convolutional layers, which may indicate several convolutional layers are needed for blending the person-specific information with image contents for PSMs prediction. Further, when the filters are inserted in all of the last three layers, the performance drops even further, possibly due to the dramatically increased parameters in CNN-PIEF and the unmatched amount of training samples.

### 5.6 The Effectiveness of Different Person-Specific Information

We also show CC, Similarity, AUC-Judd, and NSS for persons with similar/dissimilar person-specific information in Fig. 10, which are conducted based on the ground truth in our dataset. We can see that persons with similar person-specific information usually corresponds to higher consistency in terms of all saliency evaluation metrics. For example, for images in the first row of Fig. 10, we can see that persons with the same gender usually have higher CC, Similarity, AUC-Judd, and NSS. For persons with different genders, their personalized saliency maps (ground truth) exhibit heterogeneity. In Fig. 10, (e) and (f), (g) and (h) show fixation maps of different participants who like/dislike sports or fashion, respectively. The agreements among D-1 to D-5 are lower than those among L-1 to L-5, since these participants who like sports or fashion might have similar eye fixation maps, while these participants who dislike sports or fashion might have different preferences which correspond to different eye fixation maps. So this figure shows that persons with similar person-specific information usually correspond to higher consistency than that correspond to dissimilar ones. This validates the effectiveness of person-specific information for personalized saliency detection.

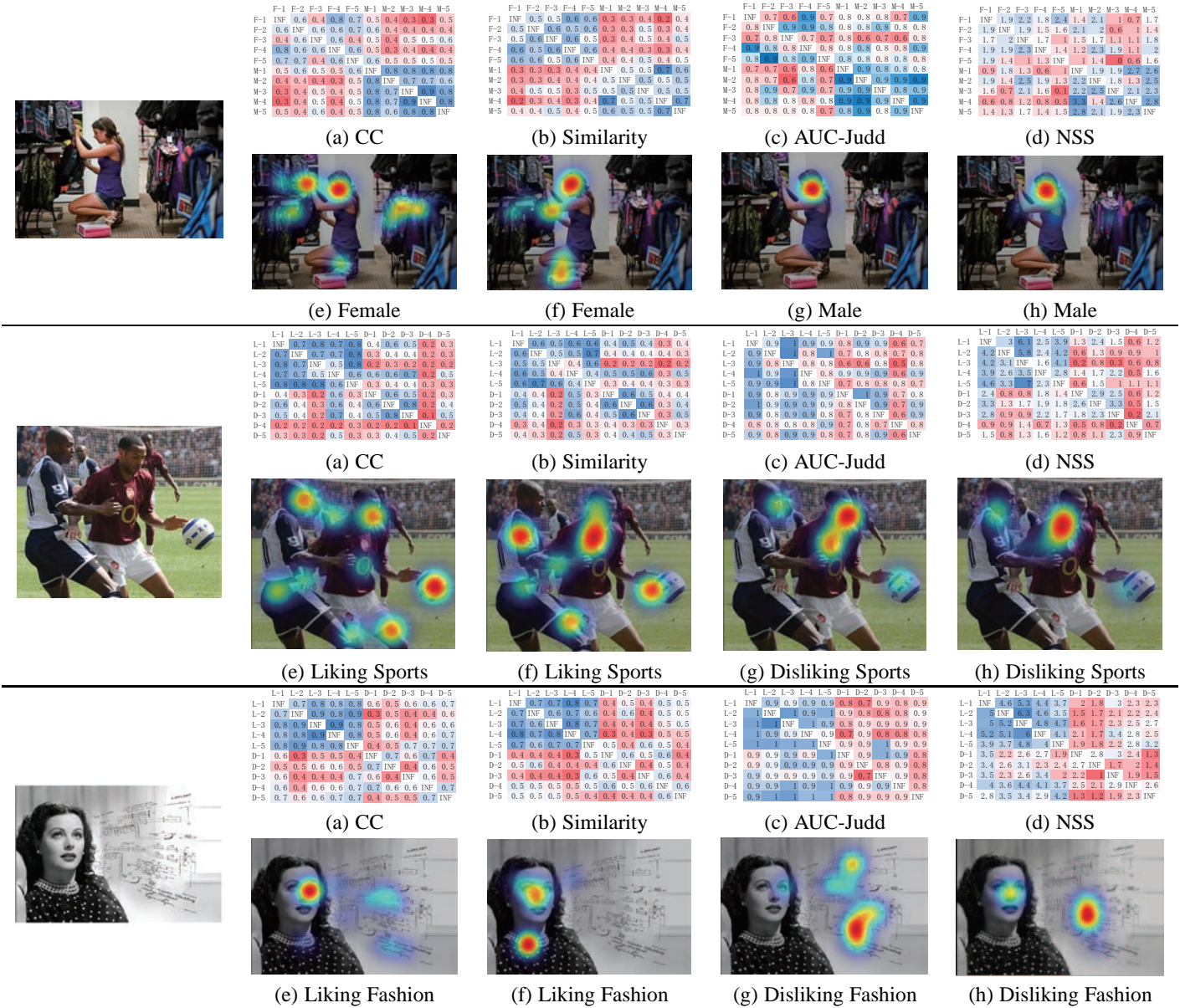

Fig. 10: The CC, Similarity, AUC-Judd, and NSS for participants with the same/different person-specific information. Here 'F/M-$x$' represents the Female/Male with index $x$, and 'L/D-$x$' represents the $x$-th participant who likes/dislikes sports or fashion, respectively. Because of space constraint, we only compare the PSMs for 10 participants with the same/different person-specific information.

### 5.7 Transfer Learning

To evaluate the transferability of USM prediction, we conduct the following experiments: we evaluate the performance of USM predicted with one method and use it as the input of another

TABLE 8: Prediction errors under different transfer learning settings.

| Methods | USM prediction method | Discrepancy prediction method | CC | Similarity | AUC-Judd | NSS |
|---|---|---|---|---|---|---|
| CNN-PIEF | LDS | LDS | 0.6532 | 0.7696 | 0.8494 | 1.6638 |
| | ML-Net | LDS | 0.6363 | 0.7717 | 0.8432 | 1.6160 |
| | ML-Net | ML-Net | 0.6368 | 0.8095 | 0.8365 | 1.5105 |
| | LDS | ML-Net | 0.5456 | 0.7784 | 0.8029 | 1.2000 |
| Multi-task CNN | LDS | LDS | 0.6509 | 0.7792 | 0.8459 | 1.6308 |
| | ML-Net | LDS | 0.6459 | 0.7634 | 0.8499 | 1.6185 |
| | ML-Net | ML-Net | 0.6463 | 0.8077 | 0.8414 | 1.4960 |
| | LDS | ML-Net | 0.5351 | 0.7810 | 0.8015 | 1.1348 |

| Insert Position | | | CC | Similarity | AUC-Judd | NSS |
|---|---|---|---|---|---|---|
| Conv5 | Conv6 | Conv7 | | | | |
| Y | / | / | **0.6532** | **0.7696** | **0.8494** | **1.6638** |
| / | Y | / | 0.6394 | 0.7709 | 0.8426 | 1.6137 |
| / | / | Y | 0.6163 | 0.7647 | 0.8348 | 1.5290 |
| Y | Y | Y | 0.5867 | 0.7622 | 0.8229 | 1.4283 |

(a)The performance of LDS based CNN-PIEF

| Insert Position | | | CC | Similarity | AUC-Judd | NSS |
|---|---|---|---|---|---|---|
| Conv5 | Conv6 | Conv7 | | | | |
| Y | / | / | **0.6368** | **0.8095** | **0.8365** | **1.5105** |
| / | Y | / | 0.6397 | 0.8082 | 0.8414 | 1.4695 |
| / | / | Y | 0.6222 | 0.8059 | 0.8343 | 1.3987 |
| Y | Y | Y | 0.5511 | 0.7919 | 0.8013 | 1.1166 |

(b)The performance of ML-Net based CNN-PIEF

| Insert Position | | | CC | Similarity | AUC-Judd | NSS |
|---|---|---|---|---|---|---|
| Conv5 | Conv6 | Conv7 | | | | |
| Y | / | / | **0.6448** | **0.7931** | **0.8486** | **1.6004** |
| / | Y | / | 0.6292 | 0.7921 | 0.8460 | 1.4857 |
| / | / | Y | 0.6289 | 0.7897 | 0.8457 | 1.4945 |
| Y | Y | Y | 0.5704 | 0.7721 | 0.8270 | 1.3024 |

(c)The performance of BMS based CNN-PIEF

| Insert Position | | | CC | Similarity | AUC-Judd | NSS |
|---|---|---|---|---|---|---|
| Conv5 | Conv6 | Conv7 | | | | |
| Y | / | / | **0.6771** | **0.7636** | **0.8588** | **1.7819** |
| / | Y | / | 0.6298 | 0.6847 | 0.8494 | 1.6157 |
| / | / | Y | 0.6724 | 0.7631 | 0.8583 | 1.7702 |
| Y | Y | Y | 0.6629 | 0.7563 | 0.8579 | 1.7422 |

(d)The performance of SalNet based CNN-PIEF

TABLE 9: The effect of inserting person-specific information at different positions in CNN-PIEF.

method for discrepancy prediction. Results are shown in Table 8. Even we use different methods to predict the USM and discrepancy, our method still achieves comparable performance, which demonstrates the transferability of our method.

## 6 CONCLUSION

Motivated by recent psychology studies that saliency is highly specific than universal, we propose to study the personalized saliency detection task. We have built the first PSM dataset. To our knowledge, this is the first comprehensive study on personalized saliency and it is expected to stimulate significant future research.

To predict PSM, the wisdom of USM motivates us to decompose PSM as the summation of USM and a discrepancy between PSM and USM. Then we propose two solutions to predict such discrepancy: i) we have presented a Multi-task CNN framework for the prediction of this discrepancy; ii) we find that personalized saliency is closely related to each observer's personal information (gender, race, major, *etc.* ). Therefore, as such discrepancy is image contents and identity related, we propose to concatenate the USM and RGB image and feed them to a CNN-PIEF to predict this discrepancy. Extensive experiments validate the effectiveness of our methods for personalized saliency prediction.

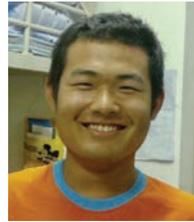

**Yanyu Xu** received the bachelor degree from Dalian University of Technology, Dalian, China. He is currently pursuing Ph.D. degree at the ShanghaiTech University.

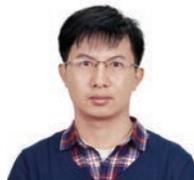

**Shenghua Gao** is an assistant professor in ShanghaiTech University, China. He received the B.Eng. degree from the University of Science and Technology of China in 2008 (outstanding graduates), and received the Ph.D. degree from the Nanyang Technological University in 2012. From Jun 2012 to Jul 2014, he worked as a postdoctoral fellow in Advanced Digital Sciences Center, Singapore. His research interests include computer vision and machine learning.

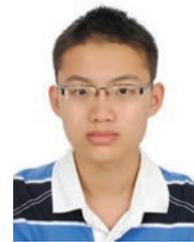

**Junru Wu** received the bachelor degree from Tongji University, Shanghai, China. From 2016 to 2017, he was a research assistant at ShngshaiTech University. He is currently pursuing Ph.D. degree at Texas A&M University.

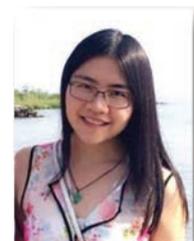

**Nianyi Li** received the BE degree from the Department of Electrical Engineering, Huazhong University of Science and Technology, in 2013. She is now working toward the PhD degree in the Department of Computer and Information Sciences, University of Delaware. Her research interests include saliency detection, object recognition, super resolution, scene reconstruction, light field rendering, and non-pinhole camera models. She is a student member of the IEEE.

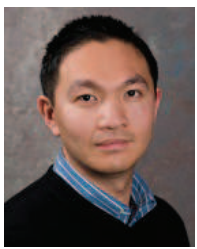

**Jingyi Yu** received the BS degree from Caltech, in 2000, and the MS and PhD degrees in EECS from MIT, in 2005. He is a professor in the Computer and Information Science Department, University of Delaware. His research interests span a range of topics in computer graphics, computer vision, and image processing, including computational photography, medical imaging, nonconventional optics and camera design, tracking and surveillance, and graphics hardware. He is a senior member of the IEEE.